\documentclass[letterpaper]{article} 
\usepackage{aaai2026}  
\usepackage{times}  
\usepackage{helvet}  
\usepackage{courier}  
\usepackage[hyphens]{url}  
\usepackage{graphicx} 
\usepackage{natbib}  
\usepackage{caption} 
\usepackage{algorithm}
\usepackage{algorithmic}
\usepackage{amsmath}
\usepackage{amssymb}
\usepackage{booktabs} 
\usepackage{multirow}
\usepackage{pifont}
\usepackage{newfloat}
\usepackage{listings}
\DeclareCaptionStyle{ruled}{labelfont=normalfont,labelsep=colon,strut=off} 
\floatstyle{ruled}
\newfloat{listing}{tb}{lst}{}
\floatname{listing}{Listing}
\title{Unsupervised Multi-View Visual Anomaly Detection via \\ Progressive Homography-Guided Alignment}
\author{
    Xintao Chen\textsuperscript{\rm1}\equalcontrib, Xiaohao Xu\textsuperscript{\rm2}\equalcontrib, Bozhong Zheng\textsuperscript{\rm 1}, Yun Liu\textsuperscript{\rm1},
    Yingna Wu\textsuperscript{\rm1}\thanks{Corresponding authors}
    }
\affiliations{
    \textsuperscript{\rm 1} ShanghaiTech University  
    \\
    \textsuperscript{\rm 2}University of Michigan, Ann Arbor
  }  
\usepackage{bibentry}

\begin{document}

\maketitle

\begin{abstract}
Unsupervised visual anomaly detection from multi-view images presents a significant challenge: distinguishing genuine defects from benign appearance variations caused by viewpoint changes. Existing methods, often designed for single-view inputs, treat multiple views as a disconnected set of images, leading to inconsistent feature representations and a high false-positive rate. To address this, we introduce ViewSense-AD (VSAD), a novel framework that learns viewpoint-invariant representations by explicitly modeling geometric consistency across views. At its core is our Multi-View Alignment Module (MVAM), which leverages homography to project and align corresponding feature regions between neighboring views. We integrate MVAM into a View-Align Latent Diffusion Model (VALDM), enabling progressive and multi-stage alignment during the denoising process. This allows the model to build a coherent and holistic understanding of the object's surface from coarse to fine scales. Furthermore, a lightweight Fusion Refiner Module (FRM) enhances the global consistency of the aligned features, suppressing noise and improving discriminative power. Anomaly detection is performed by comparing multi-level features from the diffusion model against a learned memory bank of normal prototypes. Extensive experiments on the challenging RealIAD and MANTA datasets demonstrate that VSAD sets a new state-of-the-art, significantly outperforming existing methods in pixel, view, and sample-level visual anomaly detection, proving its robustness to large viewpoint shifts and complex textures. Our code will be released to drive further research.
\end{abstract}


\section{Introduction}
\begin{figure}[t!]
    \centering
    \includegraphics[width=1\linewidth]{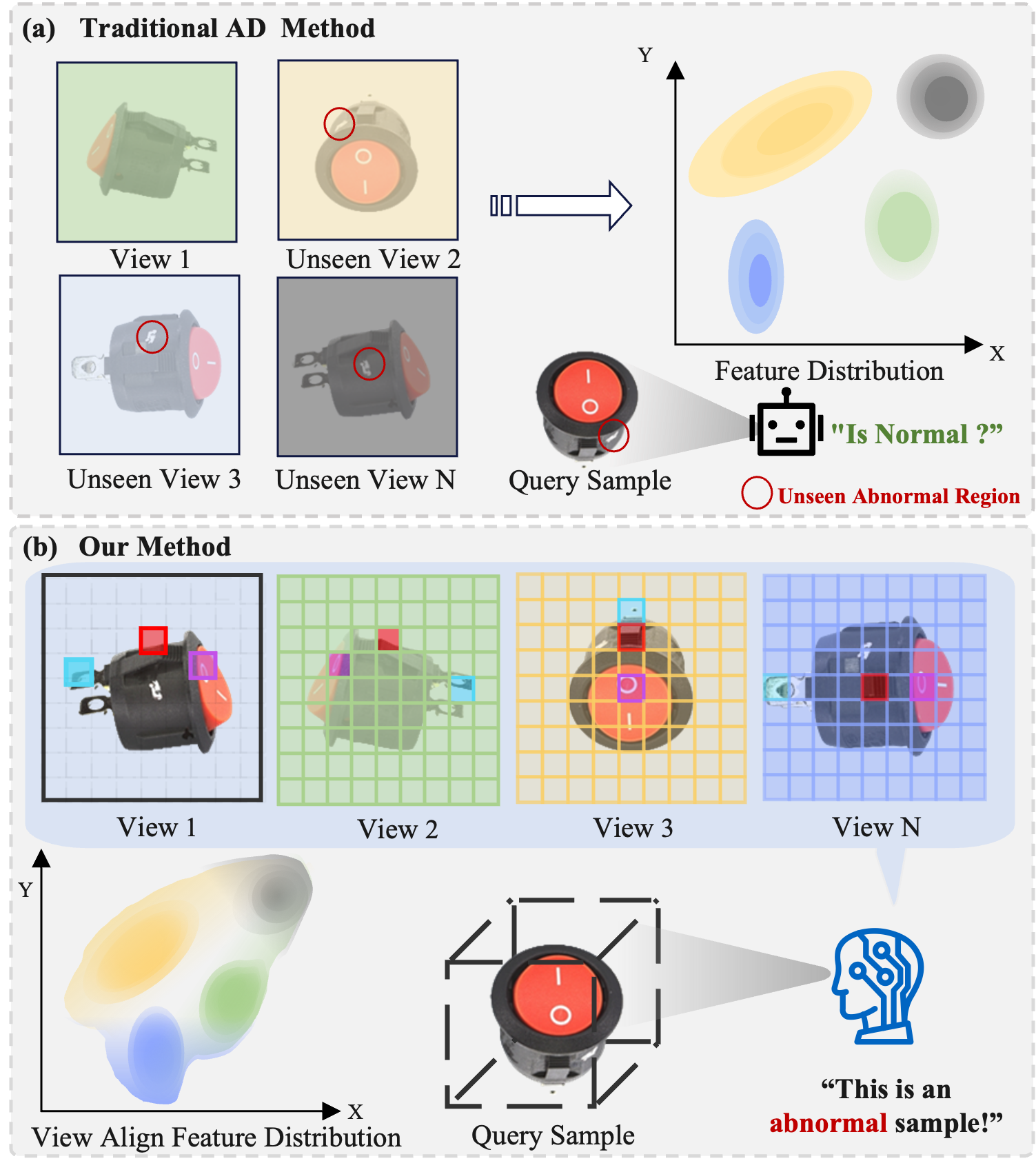}
    \caption{\textbf{(a) Conventional methods} process views independently, yielding discrete and inconsistent features that struggle to differentiate viewpoint changes from true defects. \textbf{(b) Our method (VSAD)} employs homography-based alignment to establish correspondences between views, learning a continuous and consistent representation that enables robust anomaly detection.}
    \label{fig:motivation}
\end{figure}

Industrial anomaly detection is a critical task in modern manufacturing, where even minuscule defects can compromise product quality, lead to costly recalls and pose safety risks\cite{Cao_2024}. While most unsupervised anomaly detection methods rely on single-view imagery, complex 3D objects often feature occlusions or intricate geometries that a single viewpoint cannot fully capture. Consequently, multi-view imaging systems, which capture an object from several fixed perspectives, have become a practical and effective solution for ensuring comprehensive surface inspection.

However, transitioning from single-view to multi-view settings introduces a fundamental challenge: distinguishing true anomalies from appearance shifts induced purely by changes in viewpoint (Fig.~\ref{fig:motivation}a). Existing unsupervised methods, whether reconstruction-based (e.g., DRAEM \cite{zavrtanik2021draem}) or embedding-based (e.g., PatchCore \cite{roth2022towards}), typically process each view independently. This ‘bag-of-views’ approach ignores the underlying geometric relationships between images, resulting in feature representations that are fragmented and unaligned. As a result, these models are prone to misinterpreting normal geometric variations as anomalies, leading to poor performance and reliability in real-world scenarios.

To overcome these limitations, we propose \textbf{ViewSense-AD (VSAD)}, an unsupervised framework designed to learn continuous and consistent cross-view representations (Fig.~\ref{fig:motivation}b). Our work is inspired by how human inspectors naturally operate: they mentally align different views of an object to build a holistic understanding of its surface, effortlessly distinguishing surface texture from geometric perspective shifts. VSAD mimics this reasoning process through a synergistic design. First, to determine where to look for corresponding information, we introduce a Multi-View Alignment Module (MVAM). It uses homography-based projection to explicitly match related feature patches across adjacent views. Second, to learn how to integrate this aligned information, the MVAM is embedded within a View-Align Latent Diffusion Model (VALDM). By performing alignment progressively during the denoising process, VALDM constructs a viewpoint-invariant representation. Finally, to refine the holistic alignment representation, a lightweight Fusion Refiner Module (FRM) explicitly models cross-view consistency to suppress noise and sharpen the distinction between normal and anomalous patterns.

During inference, we extract multi-level features from the DDIM inversion process and compare them against a memory bank of normal prototypes for fine-grained, multi-scale anomaly localization. Our extensive experiments on the RealIAD and MANTA datasets show that VSAD consistently outperforms state-of-the-art baselines across all evaluation levels. These results validate that by explicitly modeling geometric consistency, our framework effectively bridges the gap between fragmented image-level processing and holistic, human-like perception in multi-view anomaly detection.

Our contributions are summarized as follows:
\begin{itemize}
    \item We propose \textbf{VSAD}, a novel unsupervised multi-view anomaly detection framework that learns continuous and consistent cross-view representations through homography-guided alignment.
    \item We design a homography-based \textbf{MVAM} and embed it into a \textbf{VALDM} for progressive inter-view alignment, whose output is further enhanced by a lightweight \textbf{FRM} that refines global consistency.
    \item VSAD achieves new \textbf{state-of-the-art performance} on the large-scale RealIAD and MANTA benchmarks, demonstrating superior robustness and generalization in challenging multi-view industrial scenarios.
\end{itemize}
\section{Related Work}

\subsection{Unsupervised Anomaly Detection}
Unsupervised anomaly detection methods learn from anomaly-free data and are categorized as reconstruction-based or embedding-based. Reconstruction-based methods \cite{Fan_2025_ICCV}, like autoencoders \cite{bergmann2019mvtec}, VAE\cite{kingma2013auto} and GANs \cite{schlegl2017unsupervised}, identify anomalies as regions with high reconstruction error. More recent works have improved reconstruction fidelity using memory modules \cite{gong2019memorizing,Cai2021AppearanceMotionMC}, pseudo-anomaly augmentation \cite{zavrtanik2021draem,hu2024anomalydiffusion,sun2025unseen}, and diffusion models \cite{zavrtanik2023drak,kim2024tackling,GLAD}. However, when applied to multi-view data, they typically reconstruct each view in isolation, failing to enforce cross-view consistency. 

Embedding-based methods leverage powerful features from models pre-trained on large datasets like ImageNet \cite{deng2009imagenet}. They model the distribution of normal features using memory banks \cite{roth2022towards,bae2023pni,10.1007/978-3-031-73414-4_19}, normalizing flows \cite{gudovskiy2022cflow,yao2024hierarchical}, or student-teacher networks \cite{bergmann2020uninformed,liu2024dual,10937904}. While highly effective for single-view tasks, these methods inherently lack a mechanism to account for the geometric transformations between views, making them susceptible to feature misalignment and inconsistency in multi-view settings.

\subsection{Multi-view Feature Alignment}
Aligning features across multiple views is fundamental in computer vision, with applications in novel view synthesis \cite{gao2024cat3d,zhang2025monoinstance}, 3D perception \cite{li2022bevformer,banerjee2025hot3d}, and autonomous driving. Common strategies include Transformer-based fusion using self- or cross-attention \cite{wu2023multiview,daryani2025camuvid}, epipolar geometry constraints \cite{sun2021loftr,10387236,10949812}, and homography-based alignment \cite{he2020epinet,hwang2024booster,ni2025homer}. These techniques aim to create a unified representation by establishing spatial or semantic correspondences. However, their direct application to unsupervised anomaly detection is non-trivial. Most existing AD frameworks do not explicitly align multi-view data. Our work addresses this gap by introducing a lightweight and effective homography-based alignment mechanism tailored for industrial inspection scenarios, where objects are often captured from fixed viewpoints or on a turntable.

\subsection{Diffusion-based Models for Anomaly Detection}
Diffusion models \cite{ho2020denoising} have demonstrated state-of-the-art performance in image generation. Their ability to produce high-fidelity reconstructions has been leveraged for anomaly detection. AnoDDPM \cite{wyatt2022anoddpm} and other methods \cite{he2024diffusion,akshay2025unified} use the denoising process to reconstruct an anomaly-free version of a test image, with anomalies detected from the reconstruction residual. Others use diffusion models to synthesize diverse defects for training more discriminative models \cite{zhang2023diffusionad,hu2024anomalydiffusion,song2025defectfill}. While effective, these methods primarily focus on single-image reconstruction. In contrast, our work extracts multi-level decoder features during the DDIM \cite{song2020denoising} inversion process, not for reconstruction, but as rich descriptors for a fine-grained, embedding-based anomaly detection approach.

\begin{figure*}[t]
    \centering
    \includegraphics[width=1\linewidth]{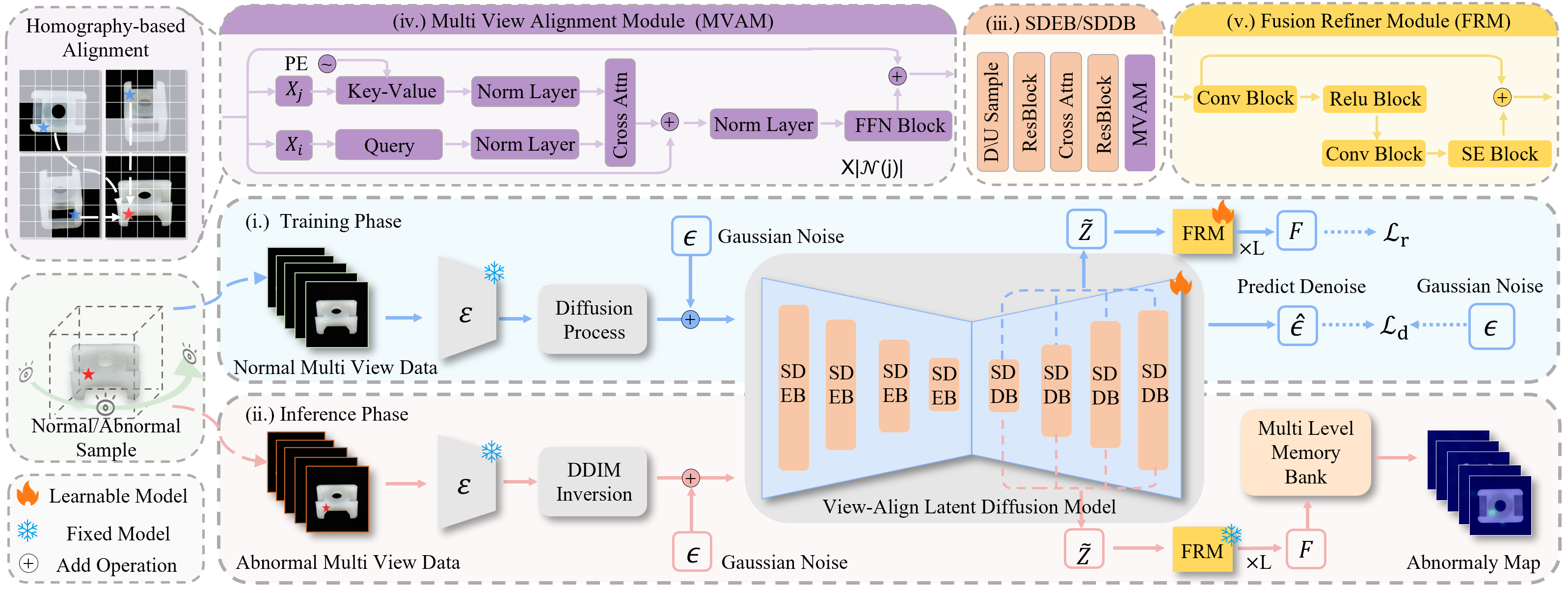}
    \caption{\textbf{Overall architecture of ViewSense-AD (VSAD)}. ({\romannumeral 1}.) During training, multi-view images are encoded into latent space. The \textbf{View-Align Latent Diffusion Model (VALDM)} performs progressive denoising, where at each Unet layer, the \textbf{Multi-View Alignment Module (MVAM)} aligns features from neighboring views using homography. A \textbf{Fusion Refiner Module (FRM)} then enhances global consistency. The model is trained with a denoising loss $\mathcal{L}_{\text{d}}$ and a refinement loss $\mathcal{L}_{\text{r}}$. ({\romannumeral 2}.) At inference, multi-level refined features are extracted via DDIM inversion and compared against a normal memory bank for anomaly scoring. ({\romannumeral 3}.) The architecture of the UNet encoder/decoder block used in Stable Diffusion. The proposed MVAM module is integrated after the ResBlock. ({\romannumeral 4}.) Detailed architecture of the MVAM. ({\romannumeral 5}.) Detailed architecture of the FRM.}
    \label{fig:pipeline}
\end{figure*}

\section{Method}
\subsection{Problem Formulation}
In unsupervised multi-view anomaly detection, we are given a training set $\mathcal{D}_C = \{S_n\}_{n=1}^N$ for an object category $C$. Each sample $S_n = \{I_m\}_{m=1}^M$ consists of $M$ RGB images captured from different viewpoints, where $I_m \in \mathbb{R}^{3 \times H \times W}$. The training set contains only anomaly-free samples. The goal is to learn a function that can identify and localize anomalies in a test sample $S_q$. This involves generating a pixel-wise anomaly map for each view, an anomaly score for each view, and an overall score for the sample.

\subsection{Overall Architecture}
We propose VSAD, an unsupervised framework designed around the principles of explicit alignment, progressive understanding, and global refinement. As illustrated in Figure~\ref{fig:pipeline}, the framework is composed of several key components. The core is a View-Align Latent Diffusion Model (VALDM) that learns the distribution of normal multi-view samples. To enable viewpoint-invariant learning, we embed our Multi-View Alignment Module (MVAM) into each layer of the model's U-Net backbone. This module uses homography to explicitly align features from neighboring views. Following alignment at each decoder stage, a lightweight Fusion Refiner Module (FRM) refines the fused features by enhancing global consistency. For detection, we use a multi-level memory bank. At inference, features are extracted from the decoder via DDIM inversion, refined, and compared against a memory bank of normal prototypes to enable robust, fine-grained anomaly scoring.

\subsection{Multi-View Alignment Module (MVAM)}
Industrial multi-view capture setups, often using fixed cameras or turntables, produce images with significant spatial overlap and smooth appearance transitions. To exploit this, we propose MVAM for patch-level feature alignment.

Given a set of multi-view feature maps $\mathbf{X} \in \mathbb{R}^{M \times C \times H \times W}$ from a U-Net layer, we define neighboring view pairs for each view $i$ as $\mathcal{P}(X_i) = \{ (X_i, X_j) \mid j \in \mathcal{N}(i) \}$, where $\mathcal{N}(i)$ is the set of adjacent view indices. For each patch centered at position $p_i$ in the query view $X_i$, we project its location into each neighboring view $X_j$ using a pre-computed homography matrix $H_{i \to j}$. Around the projected location $p_j = H_{i \to j} \cdot p_i$, we sample a local search window of size $R \times R$ to find the best-matching patch.

For each patch at location $p_j$ in the search window, we compute its relative displacement from the query patch $p_i$ after projection: $\Delta \mathbf{p}_j = p_j - H_{i \to j} \cdot p_i$. This offset is encoded using a standard 2D frequency-based positional embedding \cite{vaswani2017attention} to form $\gamma(\Delta \mathbf{p}_j)$. We then construct query, key, and value representations for attention-based aggregation:
\begin{align}
    q_i &= W_q \cdot X_i(p_i) \\
    k_j &= W_k \cdot ( X_j(p_j) + \gamma(\Delta \mathbf{p}_j) ) \\
    v_j &= W_v \cdot ( X_j(p_j) + \gamma(\Delta \mathbf{p}_j) )
\end{align}
where $W_q, W_k, W_v$ are learnable projection matrices. An attention mechanism computes weights $\alpha_j$ to aggregate the value vectors, producing an aligned feature $\tilde{X}_i(p_i)$:
\begin{equation}
\alpha_j = \frac{\exp\left( q_i^\top k_j / \sqrt{d} \right)}{\sum_{j' \in \mathcal{N}(i)} \exp\left( q_i^\top k_{j'} / \sqrt{d} \right)}
\end{equation}
\begin{equation}
\tilde{X}_i(p_i) = \sum_{j \in \mathcal{N}(i)} \alpha_j \cdot v_j
\end{equation}
This process is applied to all patches, yielding an aligned feature map $\tilde{\mathbf{X}} \in \mathbb{R}^{M \times C \times H \times W}$ with enhanced cross-view continuity.

\subsection{View-Align Latent Diffusion Model (VALDM)}
To achieve progressive understanding, we embed MVAM into a latent diffusion model based on the DDIM formulation. Given multi-view images $I=\{I_m\}_{m=1}^M$, a VAE encoder $\mathcal{E}_{\text{VAE}}$ produces latent representations $Z_0 = \mathcal{E}_{\text{VAE}}(I)$. The DDIM forward process adds Gaussian noise to produce a noisy latent $Z_t$ at timestep $t$:
\begin{equation}
    Z_t = \sqrt{\bar{\alpha}_t} Z_0 + \sqrt{1-\bar{\alpha}_t} \epsilon, \quad \epsilon \sim \mathcal{N}(0, \mathbf{I})
\end{equation}
The reverse process is handled by a U-Net which predicts the noise $\hat{\epsilon}_t$ from $Z_t$. We modify the U-Net decoder. At each decoder layer $l$, the feature map $X^{(l)}$ is first processed by the standard ResNet and attention blocks. The output is then passed to our MVAM to produce an aligned feature map $\tilde{Z}_t^{(l)}$, which is then passed to the next layer.
\begin{equation}
    \tilde{Z}_t^{(l)} = \text{MVAM}^{(l)}( \text{U-NetBlock}^{(l)}(Z_{t}^{(l)}) )
\end{equation}
This multi-stage alignment strategy allows the model to build a coherent representation by progressively aligning features at different semantic levels during the denoising process. The model is trained with a standard denoising objective:
\begin{equation}
    \mathcal{L}_{\text{d}} = \mathbb{E}_{Z_0, \epsilon, t} \left[ \|\epsilon - \hat{\epsilon}_\theta(Z_t, t)\|_2^2 \right]
\end{equation}
where $\hat{\epsilon}_\theta$ is the noise predicted by our modified U-Net.

\subsection{Fusion Refiner Module (FRM)}
While MVAM provides explicit local alignment, we introduce the FRM to further enhance global consistency and suppress noise from the fusion process. After the MVAM at each decoder layer $l$, the aligned features $\tilde{Z}^{(l)}$ are fed into the FRM. For each view $m$, FRM applies a small convolutional network $f(\cdot)$ followed by a Squeeze-and-Excitation (SE) attention block $\mathcal{A}(\cdot)$ to produce a refinement residual, which is added back to the input:
\begin{align}
    Z'^{(l)}_m &= f(\tilde{Z}^{(l)}_m) \odot \mathcal{A}(f(\tilde{Z}^{(l)}_m)) \\
    F^{(l)}_m &= \tilde{Z}^{(l)}_m + Z'^{(l)}_m
\end{align}
where $F^{(l)}_m$ is the final refined feature for view $m$ at layer $l$.

To explicitly enforce consistency, we introduce a refinement loss $\mathcal{L}_{\text{r}}$ that minimizes the L2 distance between the refined features of neighboring view pairs $(i, j)$:
\begin{equation}
    \mathcal{L}_{\text{r}} = \frac{1}{L} \sum_{l=1}^{L} \frac{1}{|\mathcal{P}|} \sum_{(i,j) \in \mathcal{P}} \| F_{i}^{(l)} - F_{j}^{(l)} \|_2^2
\end{equation}
where $L$ is the number of decoder layers and $\mathcal{P}$ is the set of all neighboring pairs. The total training loss is $\mathcal{L}_{\text{total}} = \mathcal{L}_{\text{d}} + \lambda \mathcal{L}_{\text{r}}$, where $\lambda$ is a balancing hyperparameter.

\subsection{Multi-Level Memory Bank Detection}
At inference, we use the trained model as a feature extractor. For a test sample, we perform DDIM inversion for a fixed number of steps. From each of the $L$ decoder layers, we extract the refined features $\{F^{(l)}\}_{l=1}^{L}$. During training, these features from all normal samples are stored in a multi-level memory bank $\mathcal{M} = \{\mathcal{M}^{(l)}\}_{l=1}^L$.

For a test sample, its refined features $\{F_q^{(l)}\}_{l=1}^{L}$ are extracted. The anomaly score for a patch at spatial location $(u, v)$ is calculated by finding its minimum distance to the corresponding memory bank, aggregating scores across levels:
\begin{equation}
    S_{\text{pixel}}(u,v) = \sum_{l=1}^{L} w_l \min_{m \in \mathcal{M}^{(l)}} \| F_q^{(l)}(u,v) - m \|_2
\end{equation}
where $w_l$ are weights for each level. The view-level anomaly score $S_{\text{view}}$ is the maximum score in the pixel-level anomaly map, and the sample-level score $S_{\text{sample}}$ is the maximum score across all views. This multi-level approach enables robust detection of anomalies at various scales.

\begin{table*}[t]
\centering
\setlength{\tabcolsep}{0.6mm}
\scriptsize
 {\fontsize{8}{8}\selectfont
\begin{tabular}{l|c|c|c|c|c|c|c|c|cc}

    \toprule
    \multirow{3}{*}{\textbf{Class}} & 
    \multicolumn{3}{c|}{\textbf{Reconstruction Based Method}} & \multicolumn{6}{c}{\textbf{Embeding Based Method}} \\
    \cmidrule(lr){2-10}
      & \textbf{DRAEM}& \textbf{CKAAD} & \textbf{Realnet}  & \textbf{Patchcore} & \textbf{CFlow} & \textbf{DeSTSeg} & \textbf{RDPP} & \textbf{FiCo} & \textbf{VSAD(Ours)}  \\
    \midrule

        Audiojack & 91.4/83.2/92.8 & 87.8/89.2/95.3 & 91.3/79.8/91.4 & \textbf{93.7}/81.4/97.6 & 86.5/81.1/89.9 & 88.5/81.9/95.6 & 86.4/85.4/86.3 & \underline{92.3}/89.5/\underline{98.3} & 90.3/\textbf{89.7}/\textbf{98.7} \\ 
        
        Bottle Cap & 96.3/67.6/88.1 & 95.1/92.8/97.6 & 97.3/92.8/98.5 & 94.1/91.7/94.2 & \textbf{98.9}/86.8/95.3 & 96.6/85.3/95.1 & 96.4/95.0/98.5 & \underline{98.8}/\textbf{97.7}/\underline{98.6} & 98.2/\underline{97.5}/\textbf{98.6} \\ 
        
         Button Battery & 94.6/84.4/93.1 & 91.1/85.7/96.8 & 90.6/82.5/96.6 & 81.3/79.2/96.7 & 93.3/77.8/92.0 & 92.0/\underline{90.8}/\underline{97.0} & 94.4/89.0/95.9 & 81.9/77.4/91.2 & \textbf{96.4}/\textbf{93.8}/\textbf{97.6} \\ 
         
        End Cap & 75.4/64.8/86.3 & \textbf{94.8}/\textbf{90.0}/\underline{96.4} & 81.4/72.8/92.0 & 86.7/80.6/96.2 & 83.2/75.2/89.8 & 87.2/81.5/94.0 & 93.7/86.7/96.0 & 89.0/82.7/95.9 & \underline{94.7}/\underline{89.3}/\textbf{96.8} \\ 
        
        Eraser & 69.4/70.4/81.3 & 92.3/91.9/96.3 & 88.9/87.2/98.2 & 90.6/90.7/98.3 & 92.1/90.3/\underline{98.9} & \textbf{95.6}/\textbf{96.3}/98.7 & 93.6/90.4/96.2 & 89.4/90.1/97.2 & \underline{93.8}/\underline{95.0}/\textbf{99.1} \\ 
        
        Fire Hood & 83.9/72.0/83.6 & 84.7/82.0/97.9 & 85.6/77.9/\underline{98.1} & 87.2/81.6/94.3 & 88.2/83.6/97.4 & \textbf{93.4}/\underline{88.1}/95.6 & 89.5/83.8/96.7 & 89.2/85.6/96.1 & \underline{90.4}/\textbf{88.3}/\textbf{98.4} \\ 
        
        Mint & 78.6/70.2/83.1 & 79.4/73.3/95.4 & 73.1/66.6/93.6 & 75.8/70.8/95.3 & 74.5/69.6/94.2 & 80.0/70.6/93.4 & \underline{84.7}/\underline{82.2}/\underline{95.9} & 65.5/65.9/92.6 & \textbf{85.7}/\textbf{85.8}/\textbf{97.0} \\ 
        
        Mounts & 89.3/73.3/84.0 & 97.0/85.9/98.6 & 97.2/85.0/\textbf{99.0} & \textbf{98.9}/85.4/96.4 & \underline{98.3}/85.5/98.1 & 96.1/83.4/96.9 & 95.7/\textbf{88.9}/97.1 & 96.4/83.1/96.2 & 96.9/\underline{88.5}/\underline{98.9} \\ 
        
        Pcb & 90.5/87.6/95.5 & 93.9/93.1/97.3 & 86.4/77.0/94.5 & \underline{94.5}/\underline{94.2}/97.3 & 77.4/75.8/94.3 & 91.0/87.7/97.5 & 91.8/91.9/\underline{97.7} & 89.7/89.5/96.2 & \textbf{94.6}/\textbf{94.5}/\textbf{98.2} \\ 
        
        Phone Battery & 97.8/76.6/85.4 & 93.9/\underline{91.2}/98.2 & 86.5/81.4/97.3 & 93.1/89.6/98.1 & 91.3/86.3/97.1 & 93.2/83.6/87.3 & \underline{96.4}/\textbf{91.3}/\underline{98.3} & 92.7/91.0/\textbf{98.4} & 94.3/91.1/97.5 \\ 
        
        Plastic Nut & 89.8/71.2/80.8 & 92.1/89.1/98.5 & 87.1/80.3/95.6 & 95.9/88.7/96.4 & 86.7/77.1/95.3 & 94.2/88.1/96.1 & \underline{97.0}/\underline{92.6}/\underline{98.5} & 94.0/88.9/95.9 & \textbf{97.8}/\textbf{95.2}/\textbf{98.6} \\
        
        Plastic Plug & 85.3/71.8/76.9 & 91.5/87.2/98.5 & 86.0/80.3/94.2 & 87.7/85.2/96.6 & 90.6/84.9/94.6 & 93.8/83.9/95.6 & \underline{95.3}/\underline{92.0}/\underline{98.6} & 93.7/89.2/96.1 & \textbf{95.5}/\textbf{92.1}/\textbf{99.3} \\ 
        
        Porcelain Doll & 94.0/75.7/86.2 & \underline{96.8}/86.8/97.1 & 90.2/80.1/96.9 & 90.8/79.9/96.1 & 95.2/83.4/96.3 & 94.4/81.6/95.8 & \textbf{97.2}/\textbf{90.4}/\underline{98.2} & 93.9/83.5/\textbf{98.3} & 94.6/\underline{89.3}/96.2 \\ 
        
        Regulator & 85.6/72.1/86.0 & 83.6/82.4/98.1 & 74.5/65.2/95.5 & 75.7/71.3/96.6 & 66.4/59.6/90.9 & \underline{92.2}/87.9/97.5 & 91.1/\textbf{90.1}/\textbf{98.2} & \textbf{93.8}/88.6/97.9 & 91.5/\underline{89.5}/98.1 \\ 
        
        Strip Base* & 80.0/87.8/94.9 & 99.5/97.9/99.0 & 99.0/96.8/98.1 & \textbf{99.8}/\underline{99.4}/98.8 & 99.0/97.5/98.2 & 98.9/98.4/98.8 & 99.4/99.3/\textbf{99.6} & 99.7/\underline{99.4}/\underline{99.6} & \underline{99.7}/98.9/98.6 \\ 
        
        Sim Card Set & 99.7/94.1/96.0 & 98.7/\textbf{96.8}/98.0 & 91.2/91.4/97.3 & 93.3/94.3/97.9 & 96.1/95.1/\underline{98.5} & 97.4/91.7/96.9 & 96.9/\textbf{95.9}/97.6 & 96.2/95.6/\textbf{98.7} & \underline{99.0}/94.2/97.8 \\ 
        
        Switch & 92.6/84.7/89.7 & \textbf{98.6}/95.1/97.1 & 87.7/82.7/96.1 & 94.3/93.1/96.9 & 96.0/96.0/97.9 & 96.3/95.0/99.1 & \underline{97.3}/\underline{96.7}/\underline{99.2} & 96.2/94.8/96.5 & 97.1/\textbf{96.9}/\textbf{99.3} \\
        
        Tape & 99.1/91.5/97.4 & 93.3/93.0/98.7 & 96.8/91.9/99.5 & 99.2/\underline{97.0}/98.7 & \underline{99.5}/96.6/99.2 & 98.6/94.6/99.2 & \textbf{99.8}/\textbf{97.9}/\textbf{99.6} & 99.4/96.8/99.3 & 97.4/96.8/\underline{99.6} \\
        
        Terminalblock & 68.9/55.9/83.1 & \underline{98.4}/96.3/99.4 & 92.3/82.9/91.3 & 96.6/90.5/\underline{99.6} & 95.7/88.8/97.5 & 96.8/93.1/97.5 & 93.7/\underline{96.9}/98.4 & 97.3/93.5/99.2 & \textbf{98.5}/\textbf{96.9}/\textbf{99.7} \\ 
        
        Toothbrush & 93.0/74.6/68.0 & \textbf{95.7}/\textbf{88.7}/\textbf{97.2} & 86.8/69.7/92.0 & 91.6/85.3/96.8 & 92.6/80.0/92.8 & 92.0/\underline{88.3}/95.4 & \underline{95.6}/85.3/96.9 & 90.2/83.2/95.5 & 95.3/86.7/\underline{97.0} \\
        
        Toy & 68.0/62.0/60.0 & \underline{91.8}/\underline{88.8}/95.3 & 70.8/64.0/90.2 & 91.5/82.9/\underline{95.9} & 70.1/63.2/86.4 & 91.6/82.1/89.7 & 91.5/86.6/95.7 & 84.6/78.5/90.4 & \textbf{92.6}/\textbf{88.9}/\textbf{96.1} \\ 
        
        Toy Brick & 67.6/65.7/91.2 & 79.5/77.1/92.3 & 82.4/78.1/94.2 & 79.9/69.8/92.4 & 82.1/\textbf{81.2}/\underline{96.4} & 84.8/79.0/95.3 & 82.6/78.4/95.8 & \underline{88.5}/78.4/96.1 & \textbf{89.6}/\underline{80.4}/\textbf{97.5} \\
        
        Transistor1 & 93.8/83.1/88.0 & \underline{98.7}/93.8/98.1 & 82.4/78.1/92.3 & \textbf{99.0}/94.8/98.3 & 98.1/92.6/96.9 & 97.2/95.0/96.7 & 97.8/\textbf{96.9}/97.9 & 97.6/\underline{96.3}/\underline{98.4} & 97.4/95.2/99.1 \\ 
        
        U Block & 89.7/73.9/88.6 & \underline{98.8}/\underline{92.3}/97.5 & 90.8/86.4/96.5 & 96.8/90.1/96.6 & 94.8/87.0/95.6 & 98.1/89.3/98.8 & \textbf{99.1}/\textbf{92.4}/\textbf{99.4} & 95.7/89.5/97.4 & 98.4/89.5/\underline{99.4} \\
        
        Usb & 82.2/72.6/95.7 & \textbf{95.0}/\textbf{92.7}/96.9 & 90.0/83.3/97.7 & 88.7/82.4/96.4 & 84.3/80.5/96.0 & 92.5/87.4/\underline{97.8} & 94.1/90.3/96.4 & \underline{94.5}/90.7/97.2 & 93.3/\underline{91.3}/\textbf{99.0} \\
        
        Usb Adaptor & 94.6/72.6/85.1 & 92.2/\underline{84.8}/\underline{97.2} & 84.6/72.0/95.5 & 87.0/79.9/96.8 & 86.7/80.0/94.1 & \underline{93.7}/73.9/91.5 & 92.1/82.2/96.6 & 86.0/78.6/92.0 & 93.6/\textbf{85.7}/\textbf{98.1} \\
        
        Vcpill & 82.8/75.5/75.9 & \underline{96.8}/90.7/96.7 & 92.4/88.9/\underline{98.2} & 88.7/83.5/97.3 & 91.2/89.5/97.7 & 96.6/90.6/98.1 & 96.6/\underline{91.1}/97.1 & 89.8/85.2/95.7 & \textbf{97.3}/\textbf{92.8}/\textbf{98.8} \\ 
        
        Wooden Beads & 86.8/77.8/86.1 & 90.7/\underline{87.7}/\underline{98.1} & 89.2/82.6/97.9 & 89.8/86.0/95.8 & 89.4/86.1/96.6 & \underline{92.6}/86.2/95.8 & \textbf{93.0}/\textbf{89.3}/\textbf{98.1} & 90.1/85.7/94.1 & 89.9/87.2/97.2 \\
        
        Woodstic & 76.4/72.5/90.5 & 74.3/74.5/92.6 &\textbf{92.4}/\textbf{90.6}/95.3 & 87.6/86.3/89.0 & 71.0/79.5/92.8 & 87.9/87.9/\underline{97.1} & 84.8/85.8/96.4 & 83.8/85.0/96.7 & \underline{90.6}/\underline{89.1}/\textbf{97.7} \\ 
        
        Zipper & 97.8/90.8/82.5 & \underline{99.8}/\underline{98.8}/\underline{98.8} & 99.4/94.0/97.2 & 98.8/97.9/98.1 & 98.1/95.7/96.6 & 99.4/98.5/96.2 & 84.8/85.8/98.3 & 99.8/97.5/98.0 & 99.9/98.9/99.1 \\

    \midrule
    
      \textbf{Average} & 86.5/75.9/85.9 & 92.5/89.0/97.1 & 88.1/81.4/95.7 & 90.9/86.1/96.5 & 88.9/83.5/95.3 & \underline{93.4}/87.4/96.0 & \underline{93.4}/\underline{90.0}/\underline{97.2} & 91.6/87.7/96.4 & \textbf{94.8}/\textbf{91.7}/\textbf{98.3} \\

    \bottomrule
\end{tabular} 
}
\caption{Anomaly detection performance on the \textbf{RealIAD} dataset. Scores are reported as S-AUROC / V-AUROC / P-AUROC (\%). The best result is in \textbf{bold}, second best is \underline{underlined}. ‘Scrip Base*’ denotes ‘Rolled Scrip Base’.}
\label{tab:classification_accuracy}
\end{table*}

\section{Experiments}
\subsection{Experiments Setting}
\textbf{Datasets.}
We evaluate our method on two challenging multi-view anomaly detection datasets: \textbf{Real-IAD} \cite{wang2024real} consist of 151,050 RGB images across 30 classes with 5 views, covering diverse defect types. \textbf{MANTA} \cite{fan2025manta} contains 137,338 images from 38 categories, each with 5 viewpoints. Training images are normal, while test images include both normal and defective cases.

\vspace{1mm}
\noindent\textbf{Evaluation metrics.}
We evaluate performance using the Area Under the Receiver Operating Characteristic Curve across three levels:  1) \textbf{Pixel-level AUROC (P-AUROC)} measures fine-grained anomaly localization, 2) \textbf{View-level AUROC (V-AUROC)} evaluates whether an individual view contains anomalies, and 3) \textbf{Sample-level AUROC (S-AUROC)} assesses multi-view detection by taking the maximum view-level score across all views within a sample. Higher AUROC values indicate stronger performance. More comprehensive metric comparisons are provided in the appendix of the supplementary material.

\vspace{1mm}
\noindent\textbf{Baseline methods.}
We compare our method with other state-of-the-art unsupervised approaches,including reconstruction based methods (Draem\cite{zavrtanik2021draem}, CKAAD\cite{fang2025boosting}, RealNet\cite{zhang2024realnet}) and embedding based methods (PatchCore\cite{roth2022towards}, CFlow\cite{gudovskiy2022cflow}, DeSTSeg\cite{zhang2023destseg}, RDPP\cite{tien2023revisiting}, FiCo\cite{chen2025filter}).  All results are reproduced from official code or cited from original papers.

\vspace{1mm}
\noindent\textbf{Implementation details.}
For implementation, all images are resized to $256 \times 256$. Our model is built upon the Stable Diffusion v2 implementation from the Diffusers library\cite{von-platen-etal-2022-diffusers}. We use the AdamW optimizer with a learning rate of $1 \times 10^{-4}$ and a weight decay of $1 \times 10^{-2}$. The patch sampling radius $R$ in the MVAM is set to 3. Training is conducted for 80 epochs on four NVIDIA A6000 GPUs. For the memory bank, we use features from the 3rd and 4th decoder blocks of the U-Net. More details are provided in the appendix of the supplementary material.

\begin{table*}[t]
\centering

\setlength{\tabcolsep}{0.6mm} 
 {\fontsize{8}{8}\selectfont
\begin{tabular}{l|c|c|c|c|c|c|c|c|c}
    \toprule
    \multirow{3}{*}{\textbf{Class}} & 
    \multicolumn{3}{c|}{\textbf{Reconstruction Based Method}} & \multicolumn{6}{c}{\textbf{Embedding Based Method}} \\
    \cmidrule(lr){2-10}
       & \textbf{DRAEM} & \textbf{CKAAD} & \textbf{Realnet}  & \textbf{Patchcore} & \textbf{CFlow} & \textbf{DeSTSeg} & \textbf{RDPP} & \textbf{FiCo} & \textbf{VSAD(Ours)}  \\
   \midrule
   
        Block Inductor & 85.6/81.8/71.8 & 89.5/91.2/97.2 & 76.1/77.4/88.6 & \underline{92.1}/\underline{92.5}/\underline{97.8} & 84.3/87.2/95.7 & 92.1/91.4/97.7 & 91.0/91.4/96.3 & 77.7/69.2/87.5 & \textbf{93.2}/\textbf{93.7}/\textbf{98.6} \\
        
        Copper Standoff & 66.9/61.7/71.7 & \underline{98.2}/\underline{97.9}/\underline{98.1} & 92.9/87.1/95.3 & 97.1/97.5/97.9 & 98.0/94.5/96.8 & 91.8/92.8/94.7 & \textbf{98.5}/\textbf{98.0}/\textbf{98.4} & 66.2/83.7/97.5 & 96.3/97.9/98.1 \\
        
        Flat Nut & 60.1/62.5/69.1 & 89.8/92.6/96.2 & 83.8/81.4/94.5 & \underline{95.5}/95.6/98.2 & 84.8/88.7/97.3 & 84.7/83.2/96.2 & 95.0/\underline{96.4}/\underline{98.8} & 83.0/86.2/95.9 & \textbf{96.4}/\textbf{97.7}/\textbf{99.6} \\
        
        Led & 92.0/90.1/86.9 & 97.9/97.2/98.5 & 90.8/92.6/97.3 & 96.9/97.8/98.6 & 97.9/97.8/\underline{99.1} & 96.0/92.0/97.5 & \underline{98.9}/\underline{99.0}/\textbf{99.4} & 98.8/98.2/98.6 & \textbf{99.0}/\textbf{99.8}/99.1 \\ 
        
        Led Pad & 60.7/58.8/71.7 & \textbf{98.9}/\textbf{97.2}/\textbf{98.8} & 77.1/78.3/91.8 & 98.7/97.1/97.7 & 91.4/94.0/97.1 & 98.3/\underline{97.0}/95.2 & 96.4/97.1/96.9 & \underline{98.8}/96.7/\underline{98.5} & 95.0/95.4/97.2 \\ 
        
        Long Button & 83.8/75.8/79.4 & \underline{98.2}/\underline{97.8}/\underline{98.5} & 85.5/83.9/93.1 & 96.0/96.5/98.0 & 92.1/92.5/95.8 & 95.4/97.3/98.5 & 96.3/96.9/98.1 & 95.5/96.1/97.8 & \textbf{98.3}/\textbf{97.9}/\textbf{98.6} \\ 
        
        Power Inductor & 65.1/66.3/76.9 & 85.1/86.9/95.6 & 67.1/62.9/87.4 & 81.5/82.3/94.5 & 81.1/80.7/93.7 & \textbf{87.9}/\textbf{88.7}/94.7 & 83.8/86.7/\textbf{97.0} & 80.9/83.7/\underline{96.6} & \underline{86.7}/\underline{88.3}/95.6 \\ 
        
        Short Button & 75.5/67.0/76.1 & 95.2/95.1/98.1 & 75.2/74.2/92.8 & 95.4/94.8/98.9 & 87.9/87.4/96.9 & \underline{96.2}/\underline{95.5}/\underline{99.2} & 94.7/95.3/97.0 & 93.1/92.8/98.6 & \textbf{96.9}/\textbf{96.8}/\textbf{99.4} \\
        
        Thin Resistor  & 90.1/87.8/76.2 & 97.0/\underline{97.5}/\underline{98.1} & 89.2/88.6/95.5 & 97.2/96.0/97.6 & 95.0/95.6/98.0 & \underline{98.3}/94.8/96.9 & 98.1/97.0/98.0 & 97.8/90.3/94.3 & \textbf{98.8}/\textbf{97.9}/\textbf{99.4} \\
        
        Type C & 89.3/81.9/82.7 & 96.8/97.0/98.1 & 84.0/83.7/93.3 & 97.3/95.6/96.8 & 92.6/92.8/97.7 & 97.1/97.2/98.3 & \underline{98.0}/\textbf{97.8}/\textbf{98.9} & \textbf{98.5}/\underline{97.4}/\underline{98.9} & 95.3/96.8/97.6 \\ 
        
        Wafer Resistor & 84.0/83.4/71.6 & 95.3/94.7/97.4 & 80.9/83.5/95.1 & 94.0/\underline{95.6}/96.2 & 92.4/93.0/95.3 & \underline{96.8}/94.6/96.0 & 96.1/\textbf{96.0}/\underline{99.1} & 92.5/93.2/98.7 & \textbf{97.6}/\underline{95.6}/\textbf{99.2} \\
    \midrule

        Maize & 80.8/79.8/67.8 & 83.1/82.4/87.3 & 70.7/71.2/83.2 & \textbf{85.6}/\textbf{84.1}/\textbf{92.5} & 68.6/69.7/82.1 & 84.4/81.4/79.6 & 75.6/76.6/90.7 & 61.2/67.0/89.2 & \underline{85.1}/\underline{82.9}/\underline{91.7} \\ 
        
        Paddy & 67.3/66.8/79.7 & 86.8/84.3/79.6 & \underline{88.0}/80.9/75.4 & \textbf{89.1}/\textbf{88.5}/\textbf{85.7} & 86.3/81.6/73.7 & 86.6/80.6/63.1 & 87.3/84.8/78.3 & 75.1/73.3/77.3 & 84.3/\underline{88.4}/\underline{83.2} \\ 
        
        Soybean & 78.8/83.0/68.2 & \underline{91.8}/\underline{94.4}/\underline{93.2} & 84.6/84.6/85.8 & 88.7/91.1/92.2 & 81.4/86.2/88.2 & 90.7/89.7/85.4 & 87.5/90.9/92.3 & 65.4/77.8/90.0 & \textbf{92.7}/\textbf{94.5}/\textbf{93.3} \\ 
        
        Wheat & 61.0/67.7/82.0 & 85.2/84.7/93.0 & 80.1/76.9/91.5 & 86.1/\underline{86.9}/\underline{94.6} & 85.6/86.2/94.4 & 86.0/82.7/89.0 & \underline{86.4}/85.8/93.5 & 78.7/84.7/94.1 & \textbf{86.9}/\textbf{87.6}/\textbf{96.4} \\
   \midrule
    
        Capsule & 98.8/98.0/84.6 & 97.4/96.9/87.5 & 95.7/91.9/79.8 & 98.9/98.1/93.3 & 96.5/96.8/84.7 & 98.8/98.1/89.3 & \underline{99.0}/\underline{98.4}/90.2 & 98.1/98.2/\underline{94.1} & \textbf{99.4}/\textbf{98.9}/\textbf{95.5} \\
        
        Coated Tablet & 98.6/97.6/99.0 & \underline{99.6}/\underline{99.3}/99.7 & 98.7/98.3/99.4 & 99.2/98.9/\underline{99.8} & 98.1/97.4/99.7 & \textbf{99.7}/\textbf{99.5}/\textbf{99.8} & 98.2/98.6/99.7 & 98.3/98.8/98.9 & 97.4/97.1/99.7 \\ 
        
        Embossed Tablet & 83.2/79.6/77.8 & 96.0/95.9/97.7 & 86.5/87.2/89.9 & 95.3/\underline{96.1}/97.6 & 90.0/88.9/92.3 & 95.3/94.7/93.5 & \underline{96.7}/\textbf{96.6}/\textbf{98.5} & 95.6/95.0/98.3 & \textbf{97.8}/\underline{96.1}/\underline{98.5} \\ 
        
        Lettered Tablet & 77.2/73.1/68.3 & \underline{94.9}/\underline{94.3}/\textbf{97.5} & 77.4/76.0/94.4 & 94.5/94.1/97.2 & 91.2/91.3/96.4 & 87.1/84.0/89.7 & \textbf{95.1}/\textbf{94.6}/\underline{97.3} & 92.3/93.2/96.2 & 94.3/91.4/97.0 \\ 
        
        Oblong Tablet & 72.3/70.5/76.0 & 93.2/\underline{91.6}/\underline{97.1} & 76.1/75.8/89.1 & 87.5/89.7/94.5 & 77.9/81.5/92.4 & \underline{93.5}/89.7/93.0 & 87.3/90.5/92.6 & 81.6/77.5/95.2 & \textbf{95.4}/\textbf{93.9}/\textbf{98.1} \\
        
        Pink Tablet & 94.3/91.8/92.7 & 98.3/97.5/98.3 & 95.4/94.2/98.2 & 97.7/97.3/98.2 & 96.4/95.5/\underline{98.8} & 97.4/96.5/96.9 & 98.5/97.2/98.2 & \textbf{99.0}/\textbf{98.5}/98.0 & \underline{98.8}/\underline{98.3}/\textbf{99.4} \\
        
        Red Tablet & 74.3/71.3/73.7 & \underline{82.6}/\underline{85.1}/\underline{85.0} & 77.5/75.5/69.5 & 77.1/80.6/82.9 & 77.5/79.8/71.2 & 80.0/81.4/66.4 & 79.1/82.4/84.1 & 74.9/77.6/83.6 & \textbf{87.3}/\textbf{89.6}/\textbf{89.7} \\ 
        
        White Tablet & 88.9/88.9/82.4 & \underline{98.5}/95.7/97.8 & 95.5/93.9/97.2 & 97.2/96.0/97.6 & 96.1/95.4/96.3 & \textbf{99.1}/\textbf{96.6}/\textbf{99.2} & 98.2/\underline{96.6}/\underline{98.7} & 97.9/95.9/98.2 & 98.0/96.0/98.4 \\ 
        
        Yellow Tablet & 93.8/92.7/94.5 & 98.1/97.0/97.1 & 98.3/96.5/97.6 & 98.6/\underline{98.3}/97.7 & 98.7/96.3/98.5 & \textbf{99.3}/97.4/98.4 & \underline{98.9}/\textbf{98.3}/\underline{98.9} & 98.7/95.5/98.3 & 98.4/95.0/98.6 \\ 
    \midrule
    
        Button & 92.8/94.5/96.8 & 91.0/93.8/98.5 & 92.9/88.5/99.4 & 88.7/92.5/95.2 & 76.7/86.8/93.2 & \textbf{93.9}/\underline{96.0}/99.4 & 90.1/93.8/\underline{99.5} & 89.1/82.6/92.3 & \underline{93.9}/\textbf{96.8}/\textbf{99.6} \\ 
        
        Gear & 87.7/81.6/76.3 & 96.8/96.2/98.4 & 84.8/78.2/86.3 & 96.6/95.3/\underline{99.4} & 85.1/80.4/97.1 & \underline{98.0}/\underline{96.9}/99.0 & 96.8/96.1/97.4 & 96.0/94.3/99.2 & \textbf{98.1}/\textbf{97.0}/\textbf{99.7} \\
        
        Nut & 66.5/80.2/86.8 & \underline{94.1}/92.8/97.8 & 92.7/89.6/97.6 & 93.0/87.5/96.4 & 84.2/88.4/\underline{97.9} & 82.1/80.4/94.4 & 91.0/\textbf{94.0}/\textbf{99.0} & \textbf{96.6}/92.2/97.9 & 94.0/\underline{93.2}/97.5 \\
        
        Nut Cap & 75.0/65.3/55.7 & 93.8/\underline{89.8}/\underline{97.6} & 89.4/79.7/91.4 & \underline{94.3}/89.0/97.2 & 86.2/80.3/95.0 & 87.3/77.1/93.0 & 93.0/86.3/96.1 & 92.2/87.2/96.5 & \textbf{98.8}/\textbf{92.9}/\textbf{98.4} \\ 
        
        Red Washer & 96.7/94.7/91.5 & 96.9/95.0/98.4 & 94.3/91.9/\underline{98.8} & 97.5/\underline{97.2}/98.3 & 93.0/95.0/98.6 & 96.7/95.8/97.7 & 96.6/95.9/98.2 & \textbf{99.1}/\textbf{97.6}/\textbf{99.2} & \underline{97.7}/96.6/98.6 \\ 
        
        Round Button* & 82.2/82.4/85.2 & \underline{98.9}/98.1/\textbf{99.6} & 88.1/87.9/96.7 & 97.3/97.9/98.4 & 89.9/93.0/92.8 & \textbf{99.2}/\underline{98.5}/99.2 & 97.9/98.6/99.5 & 97.2/98.0/99.5 & 95.7/96.8/98.3 \\ 
        
        Screw & 65.7/67.1/81.7 & 94.1/\underline{91.7}/95.6 & 74.5/72.2/80.6 & 90.8/88.8/94.6 & 71.1/76.6/91.8 & 90.9/88.4/96.3 & 90.6/89.0/\underline{96.4} & \underline{96.6}/89.0/\textbf{96.7} & \textbf{97.7}/\textbf{92.8}/96.4 \\
        
        Square Button* & 97.0/91.9/85.4 & 97.0/96.9/99.3 & 92.2/90.3/97.9 & \textbf{98.2}/\textbf{98.4}/\textbf{99.4} & 92.3/93.7/98.4 & 97.2/96.9/98.7 & 97.6/\underline{97.2}/\underline{99.3} & \underline{97.7}/97.1/\underline{99.3} & 95.0/95.0/98.3 \\ 
        
        Terminal & 82.2/77.3/70.7 & 96.5/95.9/97.4 & 77.0/73.7/90.7 & 93.6/94.0/96.1 & 84.1/84.5/96.5 & 96.7/95.0/98.5 & 94.0/92.8/97.2 & \textbf{97.1}/\textbf{98.2}/\textbf{98.9} & \underline{96.8}/\underline{97.6}/\underline{98.6} \\ 
        
        Wire Cap & 93.3/87.9/88.4 & 94.6/93.9/96.6 & 82.6/80.8/95.6 & 90.6/93.0/95.0 & 83.7/84.8/94.0 & \textbf{96.6}/\textbf{95.4}/98.3 & 91.0/91.1/\underline{98.5} & 88.5/92.9/98.3 & \underline{95.1}/\underline{94.2}/\textbf{98.7} \\ 
        
        Yellow Washer* & 78.5/76.5/77.9 & 93.4/92.4/95.1 & 89.6/85.7/90.4 & 92.1/91.2/\underline{95.5} & 87.5/88.3/92.8 & 92.8/91.1/94.0 & 93.3/\underline{95.1}/95.3 & \underline{96.9}/91.7/95.2 & \textbf{97.1}/\textbf{96.1}/\textbf{96.5} \\ 
   \midrule
        Coffee Beans & 65.1/60.4/62.3 & \underline{85.3}/80.9/\underline{89.6} & 67.3/61.5/70.3 & 74.9/75.5/89.5 & 74.8/72.3/84.5 & 83.6/\underline{80.9}/88.4 & 64.4/62.1/85.5 & 67.7/64.3/84.4 &\textbf{88.9}/\textbf{86.8}/\textbf{91.5} \\ 
        
        Goji Berries & 78.4/77.2/81.9 & \underline{92.4}/\underline{87.5}/\underline{93.7} & 77.7/75.1/88.6 & 89.0/85.0/91.2 & 87.8/83.7/93.5 & 86.4/79.7/87.7 & 89.2/84.3/93.1 & 92.2/81.3/90.4 & \textbf{94.2}/\textbf{89.5}/\textbf{95.3} \\ 
        
        Pistachios & 57.2/62.8/67.5 & 78.4/77.0/\underline{86.7} & \underline{79.4}/76.6/76.7 & 78.1/\underline{77.3}/83.6 & 74.7/75.9/77.8 & 61.2/60.8/69.1 & 74.6/72.9/81.5 & 71.7/67.0/78.4 & \textbf{79.8}/\textbf{78.3}/\textbf{88.1} \\ 
   \midrule
   
     \textbf{Average} & 80.0/78.4/78.7 &\underline{ 93.4}/\underline{92.8}/\underline{95.5} & 84.5/82.3/90.4 & 92.4/92.2/95.4 & 87.2/87.7/93.1 & 92.0/90.3/92.7 & 92.0/91.9/95.4 & 88.8/88.2/94.7 & \textbf{94.5}/\textbf{93.9}/\textbf{96.8} \\

    \bottomrule
\end{tabular} 
}
\caption{{Comprehensive anomaly detection results with S-AUROC / V-AUROC / I-AUROC(\%) metrics on MANTA. The best and second-best results are mark in \textbf{bold} and \underline{underlined}. ‘Round Button*’ denotes ‘Round Button Cap’, ‘Square Button*’ denotes ‘Square Button Cap’ and ‘Yellow Washer*’ denotes ‘Yellow Green Washer’.}}
\label{tab:manta_result}

\end{table*}
\subsection{Comparison with State-of-the-Art Methods}
\textbf{Quantitative results on RealIAD.}
The experimental results on the RealIAD dataset are presented in Table~\ref{tab:classification_accuracy}. Our proposed method \textbf{VSAD} achieves the highest average scores across all three metrics: \textbf{98.3\%} for P-AUROC, \textbf{91.7\%} for V-AUROC, and \textbf{94.8\%} for S-AUROC, surpassing the state-of-the-art baselines by \textbf{+1.1\%}, \textbf{+1.8\%}, and \textbf{+1.4\%}, respectively. VSAD performs particularly well in categories with large view changes (e.g., Audiojack, Zipper, PCB), indicating its ability to tell real anomalies apart from differences caused by viewpoint variations. This demonstrates that our view-alignment strategy, which helps the model learn stable and consistent features across views, effectively enhances both fine-grained localization and overall anomaly detection accuracy.

\textbf{Quantitative results on MANTA.}
Table~\ref{tab:manta_result} presents the quantitative results on the MANTA dataset, where VSAD achieves the best average performance across the dataset, with \textbf{96.8\%} in P-AUROC, \textbf{93.98\%} in V-AUROC, and \textbf{94.52\%} in S-AUROC, surpassing the state-of-the-art baseline methods by \textbf{+1.3\%}, \textbf{+1.1\%}, and \textbf{+1.1\%}. These results demonstrate the robustness and generalization ability of our model across different object categories. In challenging cases such as the “rotten core” defect in the maize category where normal and abnormal textures are very similar, VSAD ranks second among all methods. However, in categories with large viewpoint variations, such as shot button and thin register, VSAD achieves the best performance. Results from both datasets suggest that our model is more stable under viewpoint changes and gives more reliable detection results.
\begin{figure}[ht]
    \centering
    \includegraphics[width=1\linewidth]{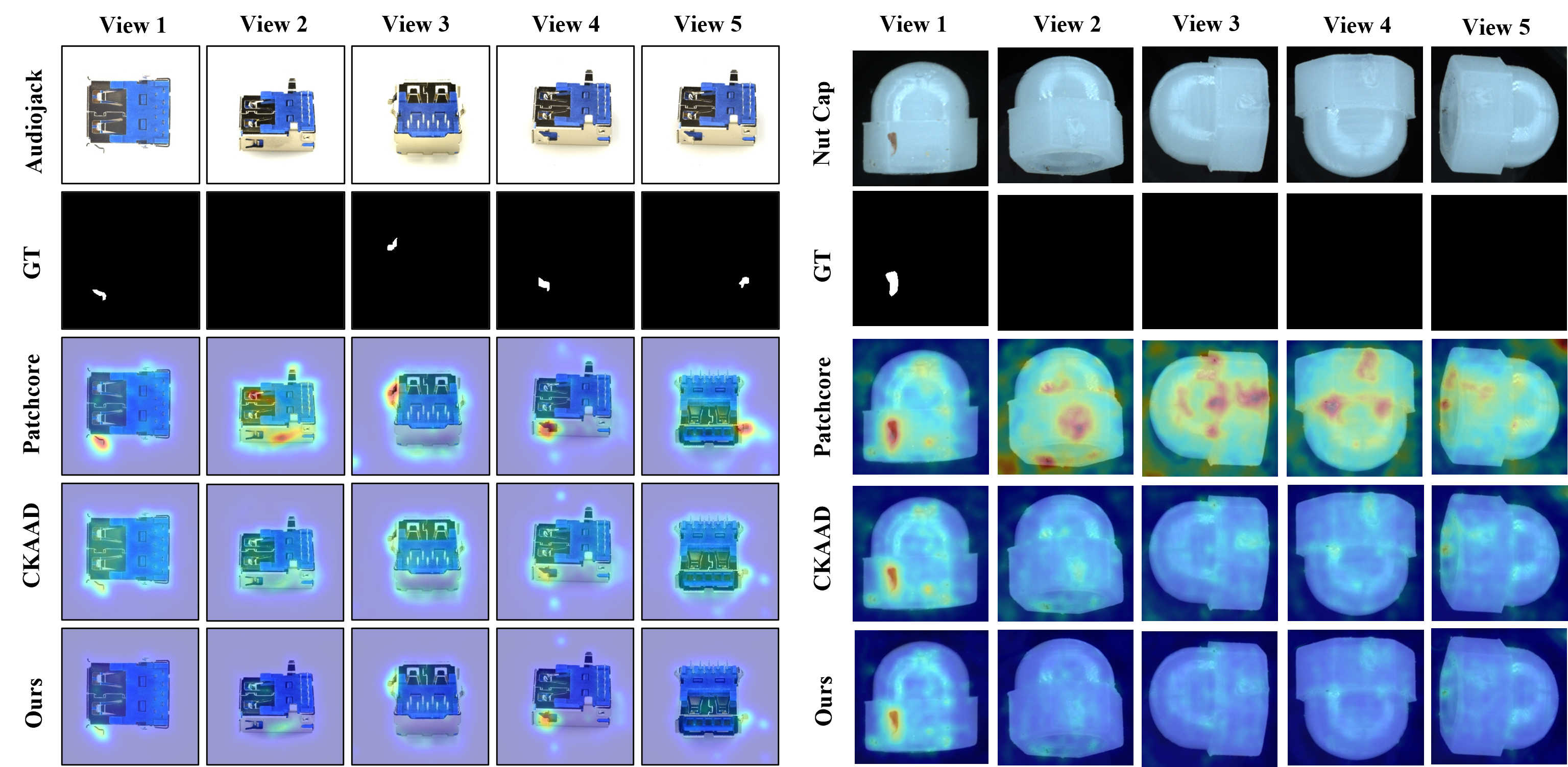}
   \caption{Qualitative anomaly localization results on RealIAD (left) and MANTA (right). Compared to baselines like PatchCore and CKAAD, our method (VSAD) produces significantly more accurate and fine-grained localization maps with fewer false positives, demonstrating its superior ability to handle viewpoint variations and subtle defects.}
   \label{fig:qualitative_results}
\end{figure}

\textbf{Qualitative results.}
To further validate the effectiveness of our model, we conduct qualitative experiments on the RealIAD and MANTA datasets. As shown in Fig. \ref{fig:qualitative_results}, our method achieves more accurate localization of anomalous regions. In comparison, the embedding based method PatchCore produces more false positives, likely due to its sensitivity to large viewpoint changes, such as the nut cap category of MANTA dataset. The reconstruction based method CKAAD performs poorly on subtle defects, especially in texture-rich objects like audiojack from RealIAD dataset, where it shows low true positive activation. Overall, our model better generalizes to viewpoint variations and complex textures, enabling finer and more precise anomaly localization in multi-view settings. Additional visualizations are included in appendix.

\subsection{Ablation Studies}
\begin{figure}[t!]
    \centering
    \includegraphics[width=1\linewidth]{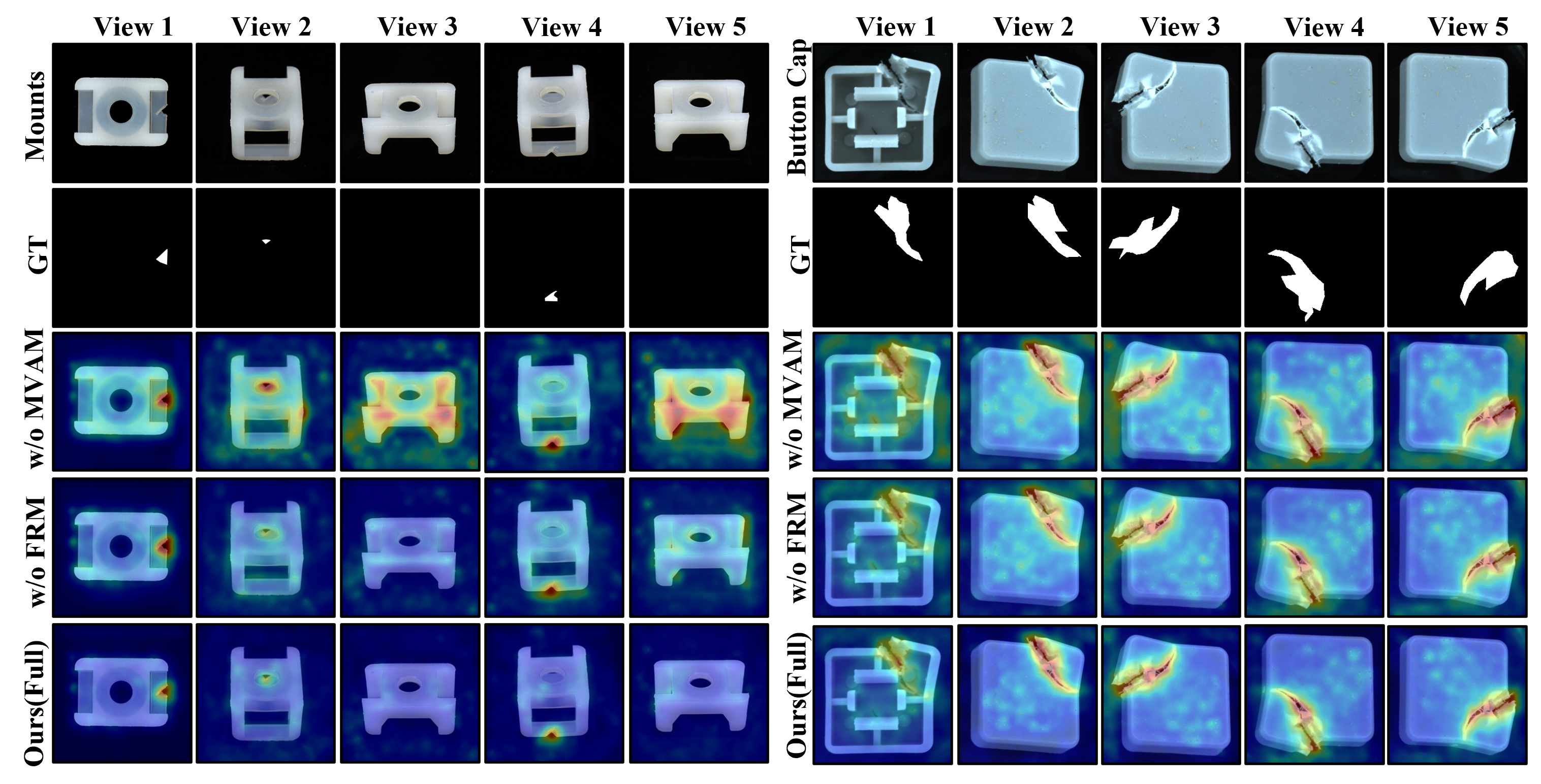}
    \caption{Visualization of the effect of component ablations on localization performance, shown on RealIAD (left) and MANTA (right).}
    \label{fig:qualitative_componnet_analysis}
\end{figure}

\textbf{Effectiveness of different components.}
As shown in Table~\ref{tab:pasdf_ablation}, we conduct ablation experiments on the RealIAD and MANTA datasets to evaluate the effectiveness of different components, with model performance assessed at the sample, view and pixel levels. We independently remove the MVAM and FRM modules from the architecture and observe consistent drops in all metrics. Specifically, removing the MVAM module causes a drop by \textbf{-8.32\%/-8.47\%/-7.09\%} reaching \textbf{86.52\%/83.24\%/91.25\%} on RealIAD dataset and on MANTA dataset by \textbf{-10.30\%/-10.38\%/-7.47\%} reaching \textbf{84.22\%/83.56\%/89.34\%}, highlighting its importance to implicitly align multi view representation. Similarly, removing the FRM modules from each decoder layer results in drops of \textbf{-1.62\%/-0.89\%/-0.99\%} on RealIAD dataset, with final scores of \textbf{93.22\%/90.82\%/97.35\%}, and \textbf{-0.90\%/-0.68\%/-0.98\%} on MANTA dataset, with scores of \textbf{93.62\%/93.26\%/95.83\%} indicating that FRM effectively refines the learned features by explicitly enforcing consistency across views and reducing noise from the fusion process for better anomaly detection. 
Fig.~\ref{fig:qualitative_componnet_analysis} presents the localization results from various component ablation studies, showing that the proposed components significantly enhance anomaly discrimination.
\begin{table}[t]
\small
\centering
 \setlength{\tabcolsep}{0.8mm}
\label{tab:component_ablation}
\resizebox{\linewidth}{!}{
\begin{tabular}{l|c|cccc}
\toprule
\textbf{Dataset} & \textbf{Method}  & \textbf{S-AUROC} ($\uparrow$) & \textbf{V-AUROC}($\uparrow$) & \textbf{P-AUROC} ($\uparrow$) \\
\midrule
\multirow{3}{*}{RealIAD} 
&w/o MVAM  &86.52&83.24&91.25  \\
&w/o FRM  &93.22&90.82&97.35     \\
&VSAD(Full) &\textbf{94.84}&\textbf{91.71}&\textbf{98.34}       \\
\midrule
\multirow{3}{*}{MANTA} 
&w/o MVAM &84.22&83.56&89.34   \\
&w/o FRM &93.62&93.26&95.83      \\
&VSAD(Full) &\textbf{94.52}&\textbf{93.94}&\textbf{96.81}       \\
\bottomrule
\end{tabular}
}
\caption{Ablation study on the key components of VSAD on the RealIAD and MANTA datasets. Performance is measured in Sample / View / Pixel-AUROC (\%).}
\label{tab:pasdf_ablation}
\end{table}
\begin{figure}[t]
    \centering
    \includegraphics[width=1\linewidth]{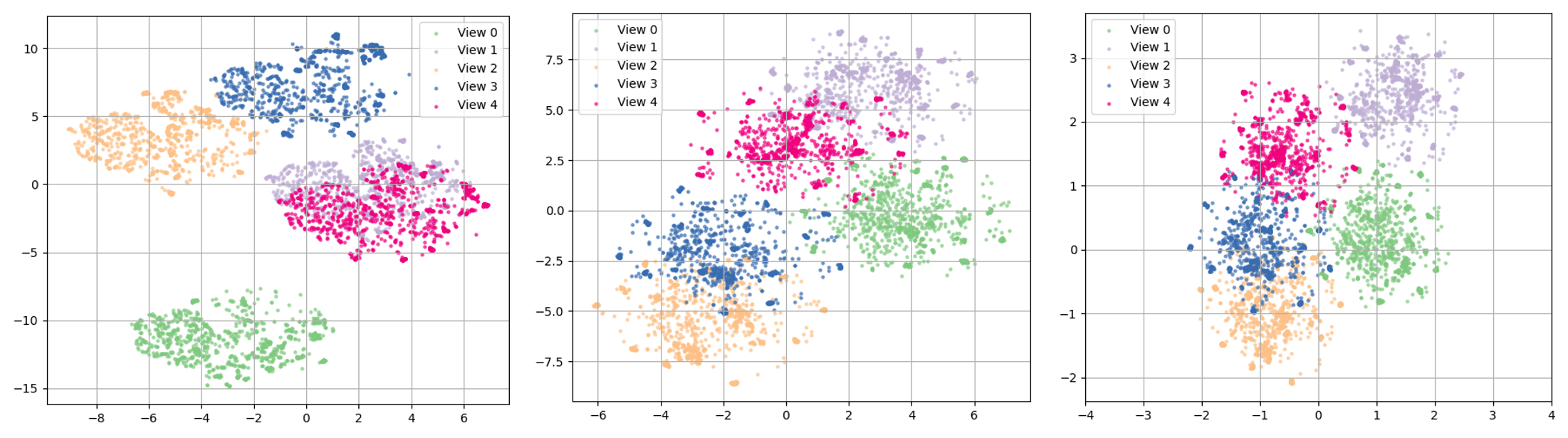}
    \caption{t-SNE visualization of multi-view features for the ‘USB’ class from RealIAD. \textbf{Left}: Before alignment, features from different views (colors) are scattered. \textbf{Middle}: After MVAM, features begin to form view-specific clusters. \textbf{Right}: After FRM, the clusters become tighter and more distinct, indicating improved feature discriminability. }
    \label{fig:tsne_vis}
\end{figure}

\begin{table}[t]

\setlength{\tabcolsep}{0.8mm}
\small
\begin{tabular}{cccc|ccc}
\toprule
\multicolumn{4}{c|}{\textbf{Layers Used}} & \multirow{3}{*}{\textbf{S-AUROC}($\uparrow$)} & \multirow{3}{*}{\textbf{V-AUROC}($\uparrow$)} & \multirow{3}{*}{\textbf{P-AUROC}($\uparrow$)} \\
\cmidrule(lr){1-4}
\textbf{4} & \textbf{3} & \textbf{2} & \textbf{1} & & & \\
\midrule
\ding{51} & \ding{55} & \ding{55} & \ding{55} & 93.7 & 91.1 & 97.8 \\
\ding{51} & \ding{51} & \ding{55} & \ding{55} & \textbf{94.8} & \textbf{91.7} & \textbf{98.3} \\
\ding{51} & \ding{51} & \ding{51} & \ding{55} & 92.3 & 90.9 & 97.4 \\
\ding{51} & \ding{51} & \ding{51} & \ding{51} & 89.4 & 86.9 & 95.5 \\
\bottomrule
\end{tabular}
\centering
\caption{Impact of using features from different U-Net decoder layers on RealIAD performance (\%). Using levels 4+3 provides the best overall performance.}
\label{tab:ablation_unet}

\end{table}

\begin{table}[t!]
\centering
\small

 \setlength{\tabcolsep}{1mm}
\begin{tabular}{c|ccc}
\toprule
\textbf{$R\times R$} & \textbf{S-AUROC}($\uparrow$) & \textbf{V-AUROC}($\uparrow$) & \textbf{P-AUROC}($\uparrow$) \\
\midrule
$ 1\times 1$  & 91.52 & 91.03 & 95.21 \\
$ 2\times 2$  & 94.32 & 93.67 & 96.15 \\
$ 3\times 3$ & \textbf{94.52} & \textbf{93.94} & \textbf{96.81} \\
$ 4\times 4$ & 93.95 & 93.54 & 95.95 \\
\bottomrule
\end{tabular}
\caption{Ablation study performance (\%) of patch sampling radius hyperparameter $R$ on the MANTA benchmark.}\label{tab:ablation_r_manta}

\end{table}

\vspace{1mm}
\noindent\textbf{Cross-view feature distribution analysis.}  As shown in Fig.~\ref{fig:tsne_vis}, before passing through the MVAM module, features from different views are more scattered in the  feature space (processed via t-SNE). After multi-stage alignment via MVAM, they form tighter clusters, which are further compacted by the FRM module. This process reduces boundary noise and improves cross-view consistency.

\vspace{1mm}
\noindent\textbf{Impact of different UNet decoder layer.}
To assess the impact of different UNet decoder levels on our embedding-based model, we conduct ablation studies using various combinations of decoder layers on both datasets. As shown in Table~\ref{tab:ablation_unet}, using features from both the 3rd and 4th decoder blocks yields the best performance on RealIAD, with improvements of  \textbf{+1.16\%}(S-AUROC), \textbf{+0.66\%}(V-AUROC), and \textbf{+0.58\%}(P-AUROC) over using only the 4th block. High-level features provide rich semantic information, while mid-level features add structural details that help improve performance. In contrast, adding lower-level features from the 2nd and 1st decoder blocks leads to performance drops, likely due to the increased noise from multi-view variations.

\vspace{1mm}
\noindent\textbf{Impact of hyperparamenters patch sampling radius.}
According to Table \ref{tab:ablation_r_manta}, an appropriate selection on hyperparamenters R greatly improves anomaly localization and detection for our method. When $R\times R=9$, the best performance of the model is achieved. We consider that a small patch sampling radius may lead to insufficient alignment between multi view representations, while increasing the radius enhances robustness and improves detection accuracy. However, when $R=4$, the model performance slightly decreases due to noise from excessive sampling.

\section{Conclusion}
We introduced ViewSense-AD, a framework tackles a key challenge in multi-view anomaly detection: distinguishing true defects from geometric variations. By embedding a homography-guided alignment module into a latent diffusion model, VSAD progressively learns viewpoint-invariant representations of object surfaces. Enhanced by a feature refinement module, this process achieves new state-of-the-art results on the RealIAD and MANTA benchmarks. Our work demonstrates that explicitly modeling cross-view geometric consistency is a robust and effective path forward for real-world industrial inspection. 

\noindent\textbf{Limitation and future work.} Future work could advance our geometric approach by replacing rigid homography with learnable deformation fields to model non-rigid objects (e.g.,textiles). Furthermore, learning the alignment transformation end-to-end would create a more flexible system that removes the reliance on pre-calibrated cameras, increasing its adaptability in dynamic settings.

\bibliography{aaai2026}

@inproceedings{zavrtanik2021draem,
  title={DRAEM--A discriminatively trained reconstruction embedding for surface anomaly detection},
  author={Zavrtanik, Vitjan and Kristan, Matej and Sko{\v{c}}aj, Danijel},
  booktitle={Proceedings of the IEEE/CVF International Conference on Computer Vision},
  pages={833--842},
  year={2021}
}

@InProceedings{Fan_2025_ICCV,
    author    = {Fan, Lei and Huang, Junjie and Di, Donglin and Su, Anyang and Song, Tianyou and Pagnucco, Maurice and Song, Yang},
    title     = {Salvaging the Overlooked: Leveraging Class-Aware Contrastive Learning for Multi-Class Anomaly Detection},
    booktitle = {Proceedings of the IEEE/CVF International Conference on Computer Vision (ICCV)},
    month     = {October},
    year      = {2025},
    pages     = {21419-21428}
}

@inproceedings{roth2022towards,
  title={Towards total recall in industrial anomaly detection},
  author={Roth, Karsten and Pemula, Latha and Zepeda, Joaquin and Sch{\"o}lkopf, Bernhard and Brox, Thomas and Gehler, Peter},
  booktitle={Proceedings of the IEEE/CVF Conference on Computer Vision and Pattern Recognition},
  pages={14318--14328},
  year={2022}
}

@inproceedings{bergmann2019mvtec,
  title={{MVTec AD--A Comprehensive Real-World Dataset for Unsupervised Anomaly Detection}},
  author={Bergmann, Paul and Fauser, Michael and Sattlegger, David and Steger, Carsten},
  booktitle={Proceedings of the IEEE/CVF conference on computer vision and pattern recognition},
  pages={9592--9600},
  year={2019}
}

@inproceedings{schlegl2017unsupervised,
  title={Unsupervised anomaly detection with generative adversarial networks to guide marker discovery},
  author={Schlegl, Thomas and Seeb{\"o}ck, Philipp and Waldstein, Sebastian M and Schmidt-Erfurth, Ursula and Langs, Georg},
  booktitle={International conference on information processing in medical imaging},
  pages={146--157},
  year={2017},
  organization={Springer}
}

@inproceedings{gong2019memorizing,
  title={Memorizing normality to detect anomaly in video: A memory-augmented autoencoder with dynamic online update},
  author={Gong, Dong and Liu, Lingqiao and Le, Vuong and Budhaditya, Saha and Rowe, Damien and Reid, Ian},
  booktitle={Proceedings of the IEEE/CVF international conference on computer vision},
  pages={1705--1714},
  year={2019}
}

@inproceedings{zavrtanik2023drak,
    title={{D-RAK}: A Denoising-based Reconstruction for Unsupervised Anomaly Kingdom Detection},
    author={Zavrtanik, Vitjan and Kristan, Matej and Sko{\v{c}}aj, Danijel},
    booktitle={Proceedings of the IEEE/CVF Winter Conference on Applications of Computer Vision},
    pages={6315--6324},
    year={2023}
}

@inproceedings{deng2009imagenet,
  title={Imagenet: A large-scale hierarchical image database},
  author={Deng, Jia and Dong, Wei and Socher, Richard and Li, Li-Jia and Li, Kai and Fei-Fei, Li},
  booktitle={2009 IEEE conference on computer vision and pattern recognition},
  pages={248--255},
  year={2009},
  organization={Ieee}
}

@inproceedings{gudovskiy2022cflow,
  title={{CFLOW-AD: Real-Time Unsupervised Anomaly Detection with Conditioned Normalizing Flows}},
  author={Gudovskiy, Denis and Ishizaka, Tomoya and Kozuka, Kazuki},
  booktitle={Proceedings of the IEEE/CVF winter conference on applications of computer vision},
  pages={1116--1126},
  year={2022}
}

@inproceedings{bergmann2020uninformed,
  title={Uninformed students: Student-teacher anomaly detection with discriminative latent embeddings},
  author={Bergmann, Paul and O'Mahony, Michael and Batzner, Kilian and Fauser, Michael and Sattlegger, David and Steger, Carsten},
  booktitle={Proceedings of the IEEE/CVF conference on computer vision and pattern recognition},
  pages={14435--14444},
  year={2020}
}

@inproceedings{li2022bevformer,
  title={{BEVFormer: Learning bird's-eye-view representation from multi-camera images via spatiotemporal transformers}},
  author={Li, Zhiqi and Wang, Wenhai and Li, Hongyang and Xie, Enze and Sima, Chonghao and Lu, Tong and Qiao, Yu and Dai, Jifeng},
  booktitle={European conference on computer vision},
  pages={1--18},
  year={2022},
  organization={Springer}
}

@inproceedings{vaswani2017attention,
  title={Attention is all you need},
  author={Vaswani, Ashish and Shazeer, Noam and Parmar, Niki and Uszkoreit, Jakob and Jones, Llion and Gomez, Aidan N and Kaiser, {\L}ukasz and Polosukhin, Illia},
  booktitle={Advances in neural information processing systems},
  volume={30},
  year={2017}
}

@inproceedings{sun2021loftr,
  title={{LoFTR: Detector-free local feature matching with transformers}},
  author={Sun, Jiaming and Shen, Zehong and Wang, Yuang and Bao, Hujun and Zhou, Xiaowei},
  booktitle={Proceedings of the IEEE/CVF conference on computer vision and pattern recognition},
  pages={8922--8931},
  year={2021}
}

@inproceedings{he2020epinet,
  title={{EPINET: A feature pyramid network for cross-view image matching}},
  author={He, Xuelian and Zhang, Yujie and Zhang, Zexin and Fang, Hong},
  booktitle={2020 International Conference on Culture-oriented Science \& Technology (ICCST)},
  pages={313--317},
  year={2020},
  organization={IEEE}
}

@inproceedings{ho2020denoising,
  title={Denoising diffusion probabilistic models},
  author={Ho, Jonathan and Jain, Ajay and Abbeel, Pieter},
  booktitle={Advances in Neural Information Processing Systems},
  volume={33},
  pages={6840--6851},
  year={2020}
}

@inproceedings{wyatt2022anoddpm,
    title={{AnoDDPM: Anomaly Detection With Denoising Diffusion Probabilistic Models}},
    author={Wyatt, Julian and Leach, Adam and Schmon, Sebastian M. and Zisserman, Andrew},
    booktitle={International Conference on Machine Learning},
    year={2022},
    organization={PMLR}
}

@article{zhang2023diffusionad,
  title={{DiffusionAD: A Generative Approach for Anomaly Detection}},
  author={Zhang, Zihan and Song, Xiang and Shen, Ziteng and You, You},
  journal={arXiv preprint arXiv:2311.16373},
  year={2023}
}

@inproceedings{song2020denoising,
  title={Denoising diffusion implicit models},
  author={Song, Jiaming and Meng, Chenlin and Ermon, Stefano},
  booktitle={International Conference on Learning Representations},
  year={2020}
}

@inproceedings{wang2024real,
  title={Real-iad: A real-world multi-view dataset for benchmarking versatile industrial anomaly detection},
  author={Wang, Chengjie and Zhu, Wenbing and Gao, Bin-Bin and Gan, Zhenye and Zhang, Jiangning and Gu, Zhihao and Qian, Shuguang and Chen, Mingang and Ma, Lizhuang},
  booktitle={Proceedings of the IEEE/CVF Conference on Computer Vision and Pattern Recognition},
  pages={22883--22892},
  year={2024}
}

@inproceedings{fan2025manta,
  title={Manta: A large-scale multi-view and visual-text anomaly detection dataset for tiny objects},
  author={Fan, Lei and Fan, Dongdong and Hu, Zhiguang and Ding, Yiwen and Di, Donglin and Yi, Kai and Pagnucco, Maurice and Song, Yang},
  booktitle={Proceedings of the Computer Vision and Pattern Recognition Conference},
  pages={25518--25527},
  year={2025}
}

@article{Cao_2024, 
title={A Survey on Visual Anomaly Detection: Challenge, Approach, and Prospect}, 
author={Cao, Yunkang and Xu, Xiaohao and Zhang, Jiangning and Cheng, Yuqi and Huang, Xiaonan and Pang, Guansong and Shen, Weiming},
journal={arXiv preprint arXiv:2401.16402},
year={2024} }

@InProceedings{GLAD,
author="Yao, Hang
and Liu, Ming
and Yin, Zhicun
and Yan, Zifei
and Hong, Xiaopeng
and Zuo, Wangmeng",
editor="Leonardis, Ale{\v{s}}
and Ricci, Elisa
and Roth, Stefan
and Russakovsky, Olga
and Sattler, Torsten
and Varol, G{\"u}l",
title="GLAD: Towards Better Reconstruction with Global and Local Adaptive Diffusion Models for Unsupervised Anomaly Detection",
booktitle="Computer Vision -- ECCV 2024",
year="2025",
publisher="Springer Nature Switzerland",
address="Cham",
pages="1--17",
}

@misc{kingma2013auto,
  title={Auto-encoding variational bayes},
  author={Kingma, Diederik P and Welling, Max and others},
  year={2013},
  publisher={Banff, Canada}
}

@inproceedings{sun2025unseen,
  title={Unseen Visual Anomaly Generation},
  author={Sun, Han and Cao, Yunkang and Dong, Hao and Fink, Olga},
  booktitle={Proceedings of the Computer Vision and Pattern Recognition Conference},
  pages={25508--25517},
  year={2025}
}

@inproceedings{Cai2021AppearanceMotionMC,
  title={Appearance-Motion Memory Consistency Network for Video Anomaly Detection},
  author={Ruichu Cai and Hao Zhang and Wen Liu and Shenghua Gao and Zhifeng Hao},
  booktitle={AAAI Conference on Artificial Intelligence},
  year={2021},
  url={https://api.semanticscholar.org/CorpusID:235306583}
}

@inproceedings{kim2024tackling,
  title={Tackling structural hallucination in image translation with local diffusion},
  author={Kim, Seunghoi and Jin, Chen and Diethe, Tom and Figini, Matteo and Tregidgo, Henry FJ and Mullokandov, Asher and Teare, Philip and Alexander, Daniel C},
  booktitle={European Conference on Computer Vision},
  pages={87--103},
  year={2024},
  organization={Springer}
}

@inproceedings{hu2024anomalydiffusion,
  title={Anomalydiffusion: Few-shot anomaly image generation with diffusion model},
  author={Hu, Teng and Zhang, Jiangning and Yi, Ran and Du, Yuzhen and Chen, Xu and Liu, Liang and Wang, Yabiao and Wang, Chengjie},
  booktitle={Proceedings of the AAAI conference on artificial intelligence},
  volume={38},
  number={8},
  pages={8526--8534},
  year={2024}
}

@inproceedings{bae2023pni,
  title={Pni: industrial anomaly detection using position and neighborhood information},
  author={Bae, Jaehyeok and Lee, Jae-Han and Kim, Seyun},
  booktitle={Proceedings of the IEEE/CVF International Conference on Computer Vision},
  pages={6373--6383},
  year={2023}
}

@inproceedings{yao2024hierarchical,
  title={Hierarchical gaussian mixture normalizing flow modeling for unified anomaly detection},
  author={Yao, Xincheng and Li, Ruoqi and Qian, Zefeng and Wang, Lu and Zhang, Chongyang},
  booktitle={European Conference on Computer Vision},
  pages={92--108},
  year={2024},
  organization={Springer}
}

@ARTICLE{10937904,
  author={Wang, Jinbao and Cheng, Jiayi and Gao, Can and Zhou, Jie and Shen, Linlin},
  journal={IEEE Transactions on Instrumentation and Measurement}, 
  title={Enhanced Fabric Defect Detection With Feature Contrast Interference Suppression}, 
  year={2025},
  volume={74},
  number={},
  pages={1-12},
  }

@inproceedings{liu2024dual,
  title={Dual-modeling decouple distillation for unsupervised anomaly detection},
  author={Liu, Xinyue and Wang, Jianyuan and Leng, Biao and Zhang, Shuo},
  booktitle={Proceedings of the 32nd ACM International Conference on Multimedia},
  pages={5035--5044},
  year={2024}
}

@InProceedings{10.1007/978-3-031-73414-4_19,
author="Liu, Chieh
and Chu, Yu-Min
and Hsieh, Ting-I
and Chen, Hwann-Tzong
and Liu, Tyng-Luh",
editor="Leonardis, Ale{\v{s}}
and Ricci, Elisa
and Roth, Stefan
and Russakovsky, Olga
and Sattler, Torsten
and Varol, G{\"u}l",
title="Learning Diffusion Models for Multi-view Anomaly Detection",
booktitle="Computer Vision -- ECCV 2024",
year="2025",
publisher="Springer Nature Switzerland",
pages="328--345",
}

@article{gao2024cat3d,
  title={Cat3d: Create anything in 3d with multi-view diffusion models},
  author={Gao, Ruiqi and Holynski, Aleksander and Henzler, Philipp and Brussee, Arthur and Martin-Brualla, Ricardo and Srinivasan, Pratul and Barron, Jonathan T and Poole, Ben},
  journal={arXiv preprint arXiv:2405.10314},
  year={2024}
}

@inproceedings{zhang2025monoinstance,
  title={MonoInstance: Enhancing monocular priors via multi-view instance alignment for neural rendering and reconstruction},
  author={Zhang, Wenyuan and Yang, Yixiao and Huang, Han and Han, Liang and Shi, Kanle and Liu, Yu-Shen and Han, Zhizhong},
  booktitle={Proceedings of the Computer Vision and Pattern Recognition Conference},
  pages={21642--21653},
  year={2025}
}

@inproceedings{banerjee2025hot3d,
  title={Hot3d: Hand and object tracking in 3d from egocentric multi-view videos},
  author={Banerjee, Prithviraj and Shkodrani, Sindi and Moulon, Pierre and Hampali, Shreyas and Han, Shangchen and Zhang, Fan and Zhang, Linguang and Fountain, Jade and Miller, Edward and Basol, Selen and others},
  booktitle={Proceedings of the Computer Vision and Pattern Recognition Conference},
  pages={7061--7071},
  year={2025}
}

@inproceedings{daryani2025camuvid,
  title={CaMuViD: Calibration-Free Multi-View Detection},
  author={Daryani, Amir Etefaghi and Bhutta, M and Hernandez, Byron and Medeiros, Henry},
  booktitle={Proceedings of the Computer Vision and Pattern Recognition Conference},
  pages={1220--1229},
  year={2025}
}

@ARTICLE{10949812,
  author={Wang, Shaoqian and Ding, Xiaokun and Mao, Yuxin and Dai, Yuchao},
  journal={Big Data Mining and Analytics}, 
  title={ETV-MVS: Robust Visibility-Aware Multi-View Stereo with Epipolar Line-Based Transformer}, 
  year={2025},
  volume={8},
  number={3},
  pages={520-533},
 }

@ARTICLE{10387236,
  author={Chang, Jiahao and He, Jianfeng and Zhang, Tianzhu and Yu, Jiyang and Wu, Feng},
  journal={IEEE Transactions on Image Processing}, 
  title={EI-MVSNet: Epipolar-Guided Multi-View Stereo Network With Interval-Aware Label}, 
  year={2024},
  volume={33},
  number={},
  pages={753-766},
 }

@article{ni2025homer,
  title={HOMER: Homography-Based Efficient Multi-view 3D Object Removal},
  author={Ni, Jingcheng and Zhao, Weiguang and Wang, Daniel and Zeng, Ziyao and You, Chenyu and Wong, Alex and Huang, Kaizhu},
  journal={arXiv preprint arXiv:2501.17636},
  year={2025}
}

@inproceedings{hwang2024booster,
  title={Booster-shot: Boosting stacked homography transformations for multiview pedestrian detection with attention},
  author={Hwang, Jinwoo and Benz, Philipp and Kim, Pete},
  booktitle={Proceedings of the IEEE/CVF Winter Conference on Applications of Computer Vision},
  pages={363--372},
  year={2024}
}

@inproceedings{wu2023multiview,
  title={Multiview compressive coding for 3D reconstruction},
  author={Wu, Chao-Yuan and Johnson, Justin and Malik, Jitendra and Feichtenhofer, Christoph and Gkioxari, Georgia},
  booktitle={Proceedings of the IEEE/CVF Conference on Computer Vision and Pattern Recognition},
  pages={9065--9075},
  year={2023}
}

@inproceedings{song2025defectfill,
  title={DefectFill: Realistic Defect Generation with Inpainting Diffusion Model for Visual Inspection},
  author={Song, Jaewoo and Park, Daemin and Baek, Kanghyun and Lee, Sangyub and Choi, Jooyoung and Kim, Eunji and Yoon, Sungroh},
  booktitle={Proceedings of the Computer Vision and Pattern Recognition Conference},
  pages={18718--18727},
  year={2025}
}

@inproceedings{akshay2025unified,
  title={A Unified Latent Schrodinger Bridge Diffusion Model for Unsupervised Anomaly Detection and Localization},
  author={Akshay, Shilhora and Narasimhan, Niveditha Lakshmi and George, Jacob and Balasubramanian, Vineeth N},
  booktitle={Proceedings of the Computer Vision and Pattern Recognition Conference},
  pages={25528--25538},
  year={2025}
}

@inproceedings{he2024diffusion,
  title={A diffusion-based framework for multi-class anomaly detection},
  author={He, Haoyang and Zhang, Jiangning and Chen, Hongxu and Chen, Xuhai and Li, Zhishan and Chen, Xu and Wang, Yabiao and Wang, Chengjie and Xie, Lei},
  booktitle={Proceedings of the AAAI conference on artificial intelligence},
  volume={38},
  number={8},
  pages={8472--8480},
  year={2024}
}

@inproceedings{zhang2023destseg,
  title={Destseg: Segmentation guided denoising student-teacher for anomaly detection},
  author={Zhang, Xuan and Li, Shiyu and Li, Xi and Huang, Ping and Shan, Jiulong and Chen, Ting},
  booktitle={Proceedings of the IEEE/CVF conference on computer vision and pattern recognition},
  pages={3914--3923},
  year={2023}
}

@inproceedings{tien2023revisiting,
  title={Revisiting reverse distillation for anomaly detection},
  author={Tien, Tran Dinh and Nguyen, Anh Tuan and Tran, Nguyen Hoang and Huy, Ta Duc and Duong, Soan and Nguyen, Chanh D Tr and Truong, Steven QH},
  booktitle={Proceedings of the IEEE/CVF conference on computer vision and pattern recognition},
  pages={24511--24520},
  year={2023}
}

@inproceedings{chen2025filter,
  title={Filter or compensate: Towards invariant representation from distribution shift for anomaly detection},
  author={Chen, Zining and Luo, Xingshuang and Wang, Weiqiu and Zhao, Zhicheng and Su, Fei and Men, Aidong},
  booktitle={Proceedings of the AAAI Conference on Artificial Intelligence},
  volume={39},
  number={3},
  pages={2420--2428},
  year={2025}
}

@inproceedings{fang2025boosting,
  title={Boosting Fine-Grained Visual Anomaly Detection with Coarse-Knowledge-Aware Adversarial Learning},
  author={Fang, Qingqing and Su, Qinliang and Lv, Wenxi and Xu, Wenchao and Yu, Jianxing},
  booktitle={Proceedings of the AAAI Conference on Artificial Intelligence},
  volume={39},
  number={16},
  pages={16532--16540},
  year={2025}
}

@inproceedings{zhang2024realnet,
  title={Realnet: A feature selection network with realistic synthetic anomaly for anomaly detection},
  author={Zhang, Ximiao and Xu, Min and Zhou, Xiuzhuang},
  booktitle={Proceedings of the IEEE/CVF conference on computer vision and pattern recognition},
  pages={16699--16708},
  year={2024}
}

@misc{von-platen-etal-2022-diffusers,
  author = {Patrick von Platen and Suraj Patil and Anton Lozhkov and Pedro Cuenca and Nathan Lambert and Kashif Rasul and Mishig Davaadorj and Dhruv Nair and Sayak Paul and William Berman and Yiyi Xu and Steven Liu and Thomas Wolf},
  title = {Diffusers: State-of-the-art diffusion models},
  year = {2022},
  publisher = {GitHub},
  journal = {GitHub repository},
  howpublished = {\url{https://github.com/huggingface/diffusers}}
}

\end{document}